\documentclass[letterpaper]{article} 
\usepackage{aaai23}  
\usepackage{times}  
\usepackage{helvet}  
\usepackage{courier}  
\usepackage[hyphens]{url}  
\usepackage{graphicx} 
\usepackage{natbib}  
\usepackage{caption} 
\usepackage{animate}
\usepackage{color}
\usepackage{booktabs}
\usepackage{multirow}
\usepackage{tabularx}
\usepackage{threeparttable}
\usepackage{appendix}
\usepackage{algorithm}
\usepackage{algorithmic}

\usepackage{newfloat}
\usepackage{listings}
\DeclareCaptionStyle{ruled}{labelfont=normalfont,labelsep=colon,strut=off} 
\floatstyle{ruled}
\newfloat{listing}{tb}{lst}{}
\floatname{listing}{Listing}
\title{AdaCM: Adaptive ColorMLP for Real-Time Universal \\ Photo-realistic Style Transfer}
\author{
    Tianwei Lin\textsuperscript{\rm 1 }\equalcontrib,
    Honglin Lin\textsuperscript{\rm 2}\equalcontrib\thanks{This work was done when Honglin Lin was an intern at Baidu.},
    Fu Li\textsuperscript{\rm 1},
    Dongliang He\textsuperscript{\rm 1},
    Wenhao Wu\textsuperscript{\rm 3, \rm 1}, \\
    Meiling Wang\textsuperscript{\rm 1}, 
    Xin Li\textsuperscript{\rm 1},
    Yong Liu\textsuperscript{\rm 2}
}
\affiliations{
    \textsuperscript{\rm 1}Baidu Inc., 
    \textsuperscript{\rm 2}Zhejiang University,
    \textsuperscript{\rm 3}The University of Sydney \\

    wzmsltw@gmail.com, linhl97@zju.edu.cn
}

\usepackage{bibentry}

\begin{document}

\maketitle

\begin{abstract}
Photo-realistic style transfer aims at migrating the artistic style from an exemplar style image to a content image, producing a result image without spatial distortions or unrealistic artifacts.
Impressive results have been achieved by recent deep models. However, deep neural network based methods are too expensive to run in real-time. 
%
Meanwhile, bilateral grid based methods are much faster but still contain artifacts like overexposure.
In this work, we propose the \textbf{Adaptive ColorMLP (AdaCM)}, an effective and efficient framework for universal photo-realistic style transfer.
First, we find the complex non-linear color mapping between input and target domain can be efficiently modeled by a small multi-layer perceptron (ColorMLP) model.
Then, in \textbf{AdaCM}, we adopt a CNN encoder to adaptively predict all parameters for the ColorMLP conditioned on each input content and style image pair.
Experimental results demonstrate that AdaCM can generate vivid and high-quality stylization results.
Meanwhile, our AdaCM is ultrafast and can process a 4K resolution image in 6ms on one V100 GPU.
\end{abstract}

\section{Introduction}

Photo-realistic style transfer is an attractive technique that migrates the artistic style of an example style image to a content image, generating an image as plausible as taken by a real camera.
However, when transferring color, existing artistic style transfer methods \cite{gatys2016image,johnson2016perceptual,park2019arbitrary} will also transfer local texture patterns from style image to content image, resulting in severe spatial distortions and unrealistic artifacts, which are not favored.
To address the spatial distortion problem in photo-realistic stylization, Luan et al. \cite{luan2017deep} introduce a  photorealism constraint on optimization-based stylization method \cite{gatys2016image}.  However, its expensive computation costs largely limit its practical applications.
To overcome this issue, Li et al. \cite{li2018closed} and Li et al. \cite{li2018learning} both adopt ``feed-forward and post-processing'' framework. 
First, PhotoWCT \cite{li2018closed} and Linear \cite{li2018learning} are proposed with relatively weak spatial distortions. Then, post-processing steps are applied to reduce remaining artifacts (optimization-based smoothing used in \cite{li2018closed} and spatial propagation network used in \cite{li2018learning}).
Yoo et al. \cite{yoo2019photorealistic} then propose an end-to-end deep model WCT$^2$, which keeps local structure via multi-scales skip connections of wavelet transform.  
However, these methods are still too slow and occupy too large GPU memory to process very high-resolution images in real-time.

Recently, Xia et al. \cite{xia2020joint} achieve real-time stylization on 4K resolution (12 Mega-pixels) images, where the mapping between pixel's input and output color is modeled as a 3$\times$4 affine transform. 
The proposed method first predicts low-resolution affine coefficient matrices in the bilateral grid, then it uses a full-resolution single-channel guidance map to slice these matrices into full-resolution coefficient matrices that are applied to each pixel of the original image.
However, the 3$\times$4 color affine transform process has limited capability especially for complex color mapping relationship, which may lead to overexposure in stylization results.
Besides,  Xia et al. \cite{xia2020joint} may cause the contrast reduction in some cases, especially along strong edges, which is a known limitation of the local affine transform model \cite{chen2016bilateral}.

%

Unlike predicting simple affine transform locally for different location in image,  we aim to model stylization with global and expressive color mapping models in this work.
\emph{What is a more capable model for color mapping?}
A widely used and expressive tool is 3D look-up table (3D-LUT), which is very fast and can describe complex color mapping relationship accurately.
However, as a large three-dimension matrix, 3D-LUTs generally need elaborate adjustments of experts, and are fixed after the adjustment. 
In essence, 3D-LUTs describe complex color mapping relationship in an exhaustive enumeration way with large parameter redundancy.
In this work, we propose to use a \textbf{Color Multi-Layer Perceptron (ColorMLP)} to model non-linear color mapping relationship and find out a small ColorMLP can accurately approximate most of 3D-LUTs with only 0.5$\%$ of their parameters.

This lightweight property leaves space for us to learn to predict parameters for ColorMLP, thus enabling adaptive color manipulation.
In this work, we propose a simple, effective and efficient framework \textbf{Adaptive ColorMLP (AdaCM)} for  photo-realistic stylization.
First, an encoder is adopted to combine context of low-resolution content and style images and predict MLP parameters.
Then, we use the ColorMLP with the predicted parameters to transfer color of each pixel in the original content image.
Furthermore, during inference, we use AdaCM to generate a 3D-LUT as intermediary, then can process  4K resolution images in real-time. 
To summarize, the main contributions are as follows:
\begin{enumerate}
\item We propose to use a lightweight and expressive ColorMLP model to approximate a fixed 3D-LUT, where a large proportion of parameters can be saved. This property enables adaptive color manipulation.
\item We introduce a novel  Adaptive ColorMLP framework for photo-realistic stylization, where per-pixel color mapping is modeled using a lightweight MLP with adaptively predicted parameters. 
\item Experiments demonstrate that our method can generate  high-quality photo-realistic stylization results without spatial distortions, meanwhile, it is efficient to process 4K resolutions images in real-time.

\end{enumerate}

\begin{figure}[t]
\centering
    \begin{minipage}{.5\textwidth}
        \includegraphics[width=.494\textwidth]{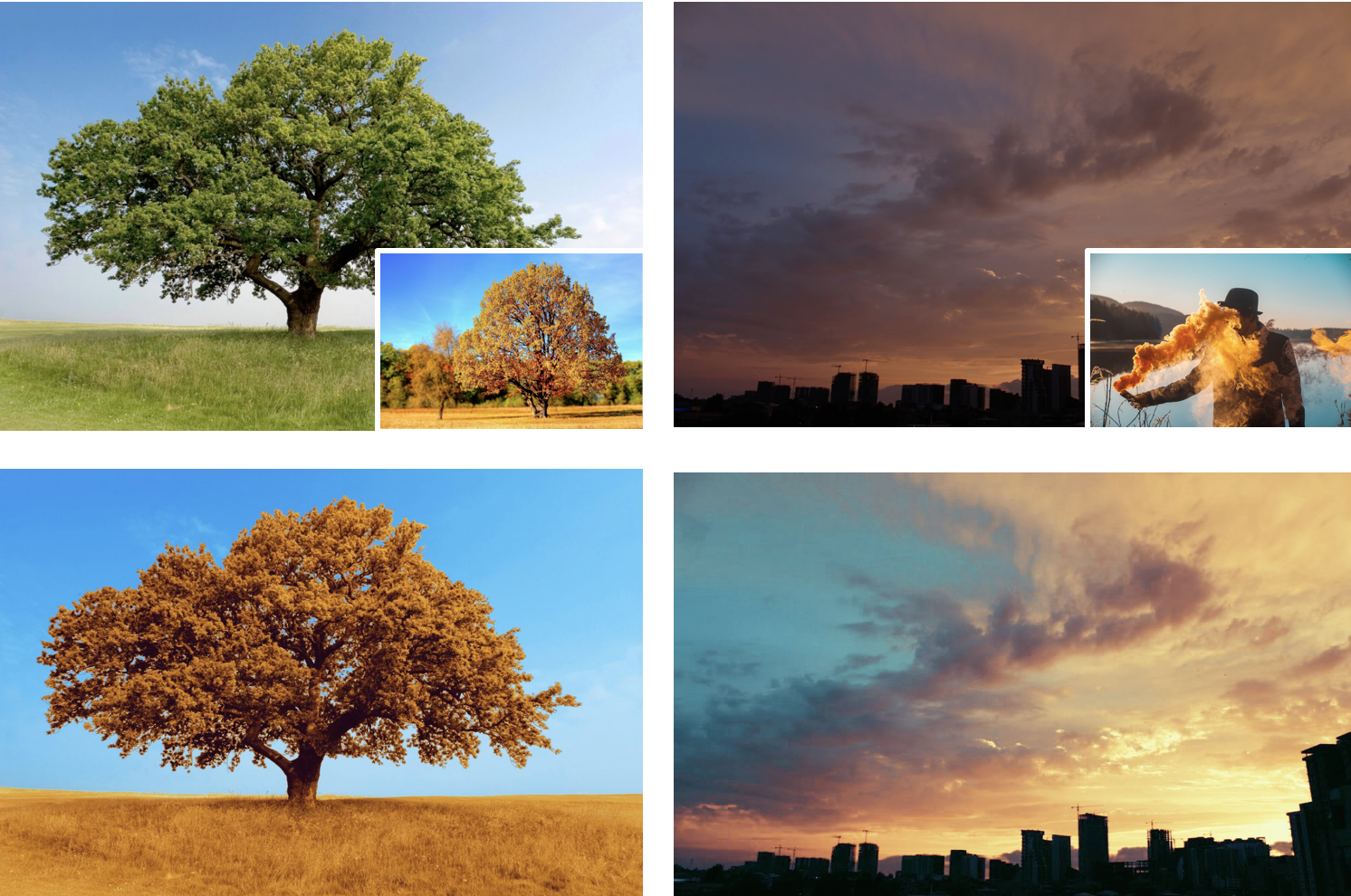}
        \animategraphics[width=.415\textwidth, autoplay, loop]{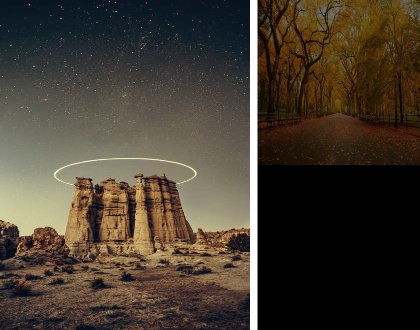}{Figures/frames/}{1}{80}
    \end{minipage}
    \caption{Illustration of our results on image/video photo-realistic style transfer. We highly recommend Adobe Acrobat to view the animated video clips at the right side.}
    \label{fig:overview}
\end{figure}

\begin{figure*}
\centering
\includegraphics[width=0.9\textwidth, height=0.25\textheight]{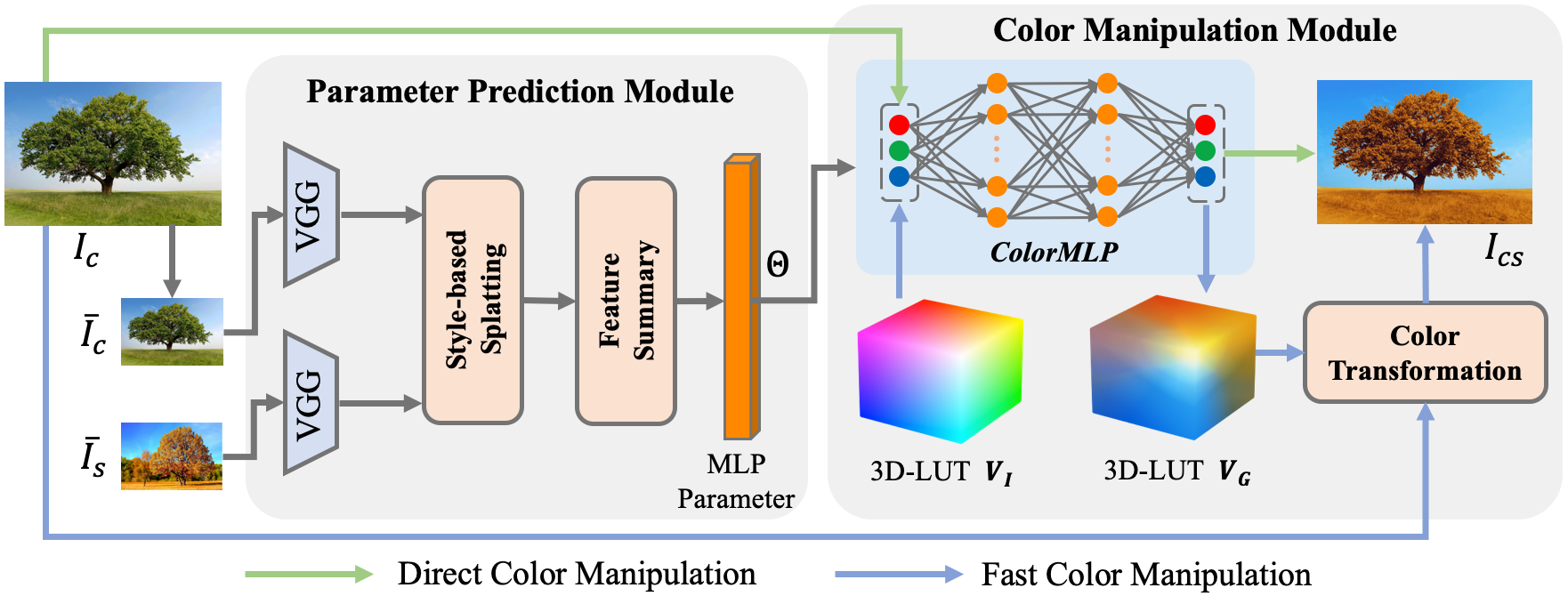} 
\caption{Overview of the full framework of AdaCM. (1) We first down-sample the content and style images to lower resolution, and feed them into the Parameter Prediction Module (PPM) to predict the parameter for ColorMLP. (2) We propose two color manipulation pipelines. The {\color{green}{``Direct''}} pipeline is used during training where each pixel is directly processed by ColorMLP. 
The {\color{blue}{``Fast''}} pipeline is used during inference, where ColorMLP is used to generate a 3D-LUT, and then this 3D-LUT is used to transfer pixel color more efficiently. }
\label{fig:framework}
\end{figure*}

\section{Related Work}

\subsection{Photo-realistic Style Transfer}
Inspired by the seminal work of Gatys et al. \cite{gatys2016image}, significant advances have been made in artistic style transfer \cite{johnson2016perceptual,li2016precomputed,huang2017arbitrary,li2017universal,sheng2018avatar,park2019arbitrary} in the past few years. While producing impressive results, these methods tend to generate severely distorted semantic structure, which is visually unpleasant in photo-realistic stylization. %
To achieve photo-realistic stylization, most early works align statistics of color distribution, including their means and variances in pixel space \cite{reinhard2001color} and histograms \cite{pitie2005n,xiao2009gradient}.
However, these global transformations ignore regional characteristics, which only transfer simple styles like color and tone. Although some methods have been proposed to model dense mapping locally \cite{shih2013data,shih2014style,wu2013content,tsai2016sky}, they are restricted to specific scenarios.

%
%

Based on Gatys' neural algorithm, Luan et al. \cite{luan2017deep} added a regularization term to obtain local affine color transformation, but it needs heavy computation to remove non-photorealistic artifacts due to the iterative optimization process. 
Li et al. \cite{li2018closed} and Li et al. \cite{li2018learning} both adopted post-processing  (optimization-based smoothing used in \cite{li2018closed} and SPN \cite{liu2017learning} used in \cite{li2018learning}) for high quality results. 
Along another line of introducing more details of high frequencies into the decoder, WCT$^2$ \cite{yoo2019photorealistic} and PhotoNAS \cite{an2020ultrafast} employed wavelet transform and skip connections respectively to  utilize the information lost during feature extraction. 
However, all these methods need to process full-resolution images with deep neural network, and the inference speed grows rapidly with increasing of resolution. Therefore, they are limited in practical applications. 
Recently, Xia et al. \cite{xia2020joint} designed joint bilateral learning method to acquire local affine transforms on each spatial grid of deep features. They achieved real-time stylization on 4K resolution (12 Mega-pixels) but sometimes leaded to blur regions due to improper coefficient matrices.

\subsection{Color Enhancement}

3D-LUTs are ultrafast operators to model color mapping relationship, and are widely used  to improve perceptual quality of digital images.
Adaptive 3D-LUT  \cite{zeng2020learning} learned 3D-LUTs from training dataset and dynamically predicted their fusion weights, 
which can better capture the joint distribution between channels compared with the curve-based transformation \cite{guo2020zero,kim2020global,kim2020pienet,song2021starenhancer} and is much more efficient than the local affine matrix \cite{gharbi2017deep}. 
However, since the basic sets of 3D-LUTs are  fixed after training, adaptive 3D-LUTs cannot achieve universal style transfer, when the color mapping relationships vary largely  for different content and style images.
Recently, Liu et al. \cite{liu2021very} enhanced the color of each pixel in image with a conditioned MLP, where the condition vector is generated by a shallow encoder.
However, the parameters of MLP are still fixed after training.
Differently, we propose to directly predict the full parameters of ColorMLP, thus can model a large variety of color mapping relationships corresponding to different input image pairs.


\section{Method}
In this section, we first give a brief description of 3D-LUT and ColorMLP, and discuss the fitting capability of ColorMLP.
Then, we introduce the proposed universal photo-realistic stylization framework AdaCM in detail.

\subsection{3D-LUT and ColorMLP}
\label{sec:3dlut}

\textbf{3D-LUT.} 
3D lookup table is an effective tool for color transfer, which can explicitly preserve complex color mapping relationship, and is widely used in many image retouching applications.
The parameters of a 3D-LUT can be denoted as $V \in R^{M\times M\times M \times 3}$, where $M$ can be regarded as the number of bins and it determines the precision of transformation. Empirically, $M$ is always set to 33 in practice.
As introduced in \cite{zeng2020learning}, the color transformation in 3D-LUT is achieved by two efficient operations: lookup and trilinear interpolation.
Take a pixel $p =\{r, g, b \}$ as input, the color transformation process of 3D-LUT can be denoted as:

\begin{equation}
p' = LUT(p, V),
\end{equation}

\noindent
\textbf{ColorMLP.} 
In this work, we explore to utilize MLP to model complex color mapping with fewer parameters.
As shown in Fig. \ref{fig:framework}, the proposed ColorMLP model has two hidden layers, each contains $N$ units.
Thus, the weight and bias parameters of a ColorMLP can be denoted as a vector $\Theta \in R^{D}$, where $D=N^2 + 8 N + 3$. And the corresponding color transformation process is:

\begin{equation}
p' = ColorMLP(p, \Theta).
\end{equation}

\begin{figure}[t]
\centering
\includegraphics[width=0.8\linewidth]{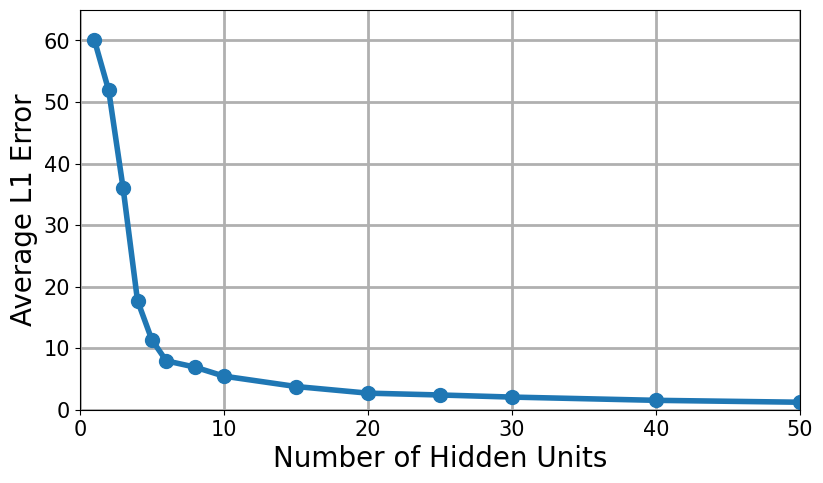}
\setlength{\belowcaptionskip}{-0.5cm}
\caption{Average L1 error \emph{versus} number of hidden units in MLP. 
}
\label{fig:fitting_exp}
\end{figure}

\noindent
\textbf{Fitting Capability of ColorMLP.}
Can we use ColorMLP to accurately approximate a given 3D-LUT?
To achieve this, we first randomly sample a large amount of colors as input and obtain target colors using 3D-LUT. Then, we use these generated color pairs to train our ColorMLP model.
After  convergence, we uniformly sample $M^3$ colors, and calculate the average L1 error between colors transferred by ColorMLP and 3D-LUT.
We conduct experiments on 10 collected 3D-LUTs of different styles, and take average error as results.
For each 3D-LUT, we train ColorMLP with different number of hidden units.
The experimental results are shown in Fig. \ref{fig:fitting_exp}, which demonstrates that larger $N$ can bring lower error.
We set the pixel value range as  0$\sim$255 during evaluation. When $N=20$, the pixel-wise L1 error is only 2.69, which is already hard for human to distinguish the differences.
Thus, a two-layer ColorMLP with $N=20$ units is enough to fit a 3D-LUT.
This lightweight property ($0.5\%$ parameters compared to 3D-LUT) leaves us space to further devise adaptive color transformation.

\subsection{Adaptive ColorMLP}

Based on the lightweight property of ColorMLP, we propose the Adaptive ColorMLP framework to achieve universal photo-realistic style transfer.
%
As shown in Fig. \ref{fig:framework}, our proposed AdaCM takes a content image $I_c$ and an arbitrary style image $I_s$ as inputs, and eventually synthesizes a stylized image $I_{cs}$.
To achieve this, at first, we down-sample the input images to low resolution counterparts $\bar{I}_c$ and $\bar{I}_s$.
Then, a \emph{Parameter Prediction Module (PPM)} will combine the feature statistics of $\bar{I}_c$ and $\bar{I}_s$, and predicts the parameters $\Theta$ for ColorMLP.  This process can be formulated as:

\begin{equation}
\Theta = PPM(\bar{I}_c, \bar{I}_s),
\end{equation}

In the final step,  the color of original content image $I_c$ is transferred based on the predicted parameter $\Theta$ and generates stylized image $I_{cs}$. 

\subsection{Parameter Prediction Module}

The Parameter Prediction Module (PPM) aims at modeling the color mapping relationship between content and style images via predicting the parameters of ColorMLP. 
The core is to match feature statistics of content image and style image.
To achieve this, following \cite{xia2020joint}, (1) we first adopt pre-trained VGG-19 \cite{simonyan2015very} network to extract multi-resolution feature maps of content and style images; (2) then we use multiple \emph{Style-based Splatting}  \cite{xia2020joint} blocks to fuse content and style feature maps based on  AdaIN \cite{huang2017arbitrary}; (3) finally we adopt multiple convolution and fully-connection layers to summarize feature statistics and predict parameters $\Theta$ of ColorMLP. 
In this way, we can fully fuse the content and style statistic in multiple feature granularity. Meanwhile, since PPM processes images in low-resolution, its computation cost is limited. More implementation details of PPM are provided in the supplementary material.

\subsection{Color Manipulation Module}

As shown in Fig. \ref{fig:framework}, the direct way of color manipulation is to directly use ColorMLP with predicted parameters to process the color of each pixel. 
However, with the growth of resolution, there are too many pixels in an image. For example, a 4K resolution image contains around 12 Mega-pixels.
Thus, although ColorMLP is extremely lightweight, the overall computation cost will still grow obviously with the increasing resolution.
%
%
%
Meanwhile, 3D-LUT is ultrafast even under high resolution due to its $O(1)$ complexity and $0$ float-point operation in need for each input pixel.
Thus, to further improve the inference efficiency, we propose a fast inference pipeline with the help of 3D-LUT.
As shown in Fig. \ref{fig:framework}, we first use ColorMLP to generate an equivalent 3D-LUT as:

\begin{equation}
V_G = ColorMLP(V_I, \Theta ),
\end{equation}
where $V_I$ is the parameters of identity 3D-LUT (each color maps to itself), and $V_G$ is the parameters of generated 3D-LUT. Then, we conduct color transformation as:

\begin{equation}
I_{cs} = LUT(I_c, V_G ).
\end{equation}

As shown in Table. \ref{table:efficiency}, with the growth of image resolution, this 3D-LUT based pipeline is extremely faster than directly transferring image color using ColorMLP.
In addition, via generating $V_G$, we can also intuitively visualize the color mapping relationship modeled by ColorMLP.

\subsection{Loss Function}

The proposed AdaCM is trained end-to-end since it is fully differentiable. Our overall loss function is the weighted summation of content loss $\mathcal{L}_c$, style loss $\mathcal{L}_s$ and the regularization term $R$:

\begin{equation}
\mathcal{L}_{all} = \lambda_c \mathcal{L}_c + \lambda_s \mathcal{L}_s + \lambda_r \mathcal{R}_{all},
\end{equation}
where $\lambda_c$, $\lambda_s$ and $\lambda_r $ are hyper-parameters controlling weights of their corresponding loss terms. 
Details of each term will be explained in the remaining part of this section.

\noindent
\textbf{Content and Style Losses.} 
For content loss, following previous works \cite{huang2017arbitrary,park2019arbitrary}, we adopt the commonly used normalized perceptual loss.
The content loss is defined as:

\begin{equation}
\mathcal{L}_c = \sum_{i=1}^4 \left \| norm(F_c^i) - norm(F_{cs}^i)  \right \| _2,
\end{equation}
where $F^i$ denotes the $i$-th layer of feature map extracted by pre-trained VGG model, and $norm$ denotes the channel-wise normalization.
For style loss, we  adopt the commonly used  mean-variance loss \cite{huang2017arbitrary} as:

\begin{equation}
\mathcal{L}_s = \sum_{i=1}^3 \left (  \left \| \mu (F_s^i) - \mu (F_{cs}^i)  \right \| _2 + \left \| \sigma (F_s^i) - \sigma (F_{cs}^i)  \right \|_2 \right ) ,
\end{equation}
where $\mu$ and $\sigma$ calculate the mean and variance of the feature vectors separately.

\noindent
\textbf{Regularization.} 
Using content and style losses, we can train the AdaCM to achieve universal photo-realistic stylization.
However, as shown in Fig. \ref{fig:exp_ablation_loss}, generated results may contain banding artifacts or abrupt color changes in original smooth areas.
%
indicating that smoothness is not carefully considered during stylization.
To ensure the generated ColorMLP containing smooth color mapping, we first consider the total variation loss $\mathcal{R}_{lut}$ on 3D-LUT as proposed in \cite{zeng2020learning}, which penalizes differences between adjacent cells of the generated 3D-LUT. 
To achieve this, we first generate a 3D-LUT $V_G$ using ColorMLP, and conduct the regularization $\mathcal{R}_{lut}$ on 3D-LUT $V_G$.
%
To be specific, we denote $v = (i, j, k)$ as one point in $V_G$ and its color is $V_G(v)$. Then,  $\mathcal{R}_{lut}$ can be defined as:

\begin{equation}
\mathcal{R}_{lut} = \sum_{v \in V_G} \sum_{v_n \in \Omega }  \left \|V_G(v) - V_G(v_n) \right \|_2  ,
\end{equation}
where $\Omega = \{(i+1, j,k), (i, j+1, k), (i, j, k+1) \}$ denotes the neighbors of $v$.
As shown in Fig. \ref{fig:exp_ablation_loss}, artifacts are largely reduced but still exist in some scenarios.
The main reason is that $\mathcal{R}_{lut}$ smooths the local mapping relationship around all colors in color space, while the color distribution of a usual content image lies only a small part of color space.
Thus, we further adopt a regular total variance loss $\mathcal{R}_{img}$ on image $I_{cs}$, which focuses on the local mapping smoothness around colors appeared in content image rather than all colors.
The overall regularization term is defined as:

\begin{equation}
\mathcal{R}_{all} = \mathcal{R}_{lut} + \mathcal{R}_{img}
\end{equation}

%

%
%

%
%

\section{Experiment}

\subsection{Dataset and Setup}

\textbf{Dataset.} 
We combine all images from Landscape \cite{afifi2021histogan} dataset and 6000 images randomly sampled from MS-COCO \cite{lin2014microsoft} dataset as our training set. During training, we randomly sample both content images and style images from this mixed dataset. 
During evaluation, we use high-quality test images collected by \cite{xia2020joint}.
%
Some copyright free images from Pexels\footnote{www.pexels.com} are also used during testing.

\begin{table*}[t]
\begin{center}
\begin{tabular}{ccccccccc}
\toprule
Method & Reinhard & Piti\'{e} & PhotoWCT & 
Linear & WCT$^2$ & PhotoNAS & Bilateral &
AdaCM \\
\midrule
SSIM-Edge $\uparrow$ & 0.7631 & 0.7443 & 0.6315 & 0.7098 & 0.6758 & 0.7057 & 0.7765 &  0.7224 \\
SSIM-Whole $\uparrow$ & 0.7746 & 0.7702 & 0.6623 & 0.7253 & 0.5987 & 0.7123 & 0.7766 &  0.7478 \\
LPIPS $\downarrow$ & 0.1419 & 0.1556 & 0.2993 & 0.2360 & 0.2255 & 0.2135 & 0.1553 &  0.1780 \\
Gram Loss $\downarrow$ & 8.0799 & 3.9796 & 3.6637 & 3.6781 & 3.6227 & 6.6976 & 3.3856 &  \textbf{3.1613} \\
\bottomrule
\end{tabular}
\caption{Quantitative evaluation results.}
\label{table:metric}
\end{center}
\end{table*}

\subsection{Qualitative Results}
\label{sec:exp_qualitative}

In this section, we compare our proposed method against three state-of-the-art photo-realistic style transfer methods: Linear \cite{li2018learning}, WCT$^2$ \cite{yoo2019photorealistic}, and Bilateral \cite{xia2020joint} using default settings.
Note that for Bilateral \cite{xia2020joint}, we directly use stylization results provided by authors, since there are no available open-source codes.
Given the limitation of length, we present more comparisons with PhotoWCT \cite{li2018closed} and PhotoNAS \cite{an2020ultrafast} in the supplementary material.

\begin{figure}[t]
\begin{center}
\includegraphics[width=0.9\linewidth]{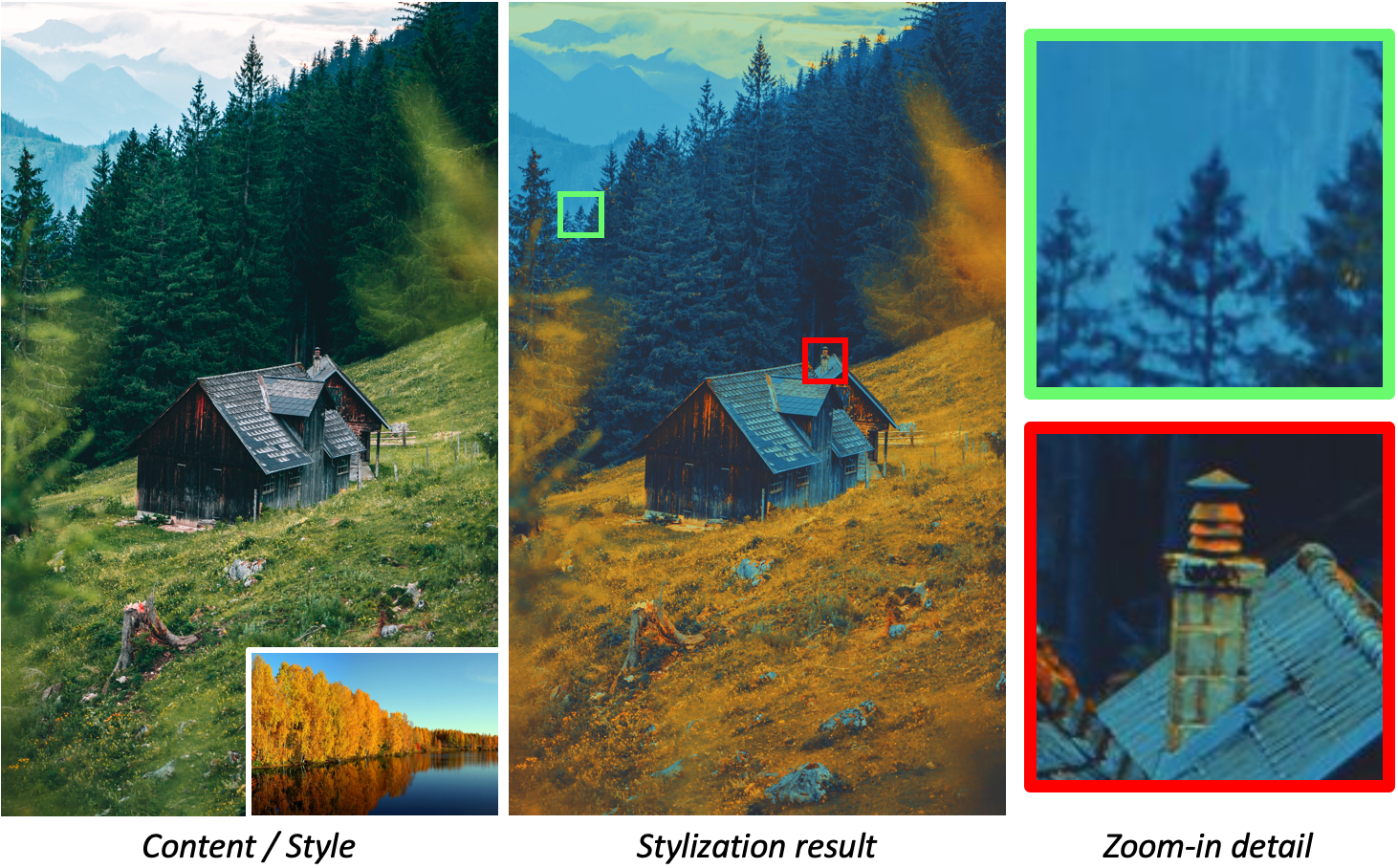}
\end{center}
   \caption{Visualization of 4K-resolution stylization. }
\label{fig:exp_high}
\end{figure}

\begin{figure*}[t]
\centering
\includegraphics[width=0.7\linewidth]{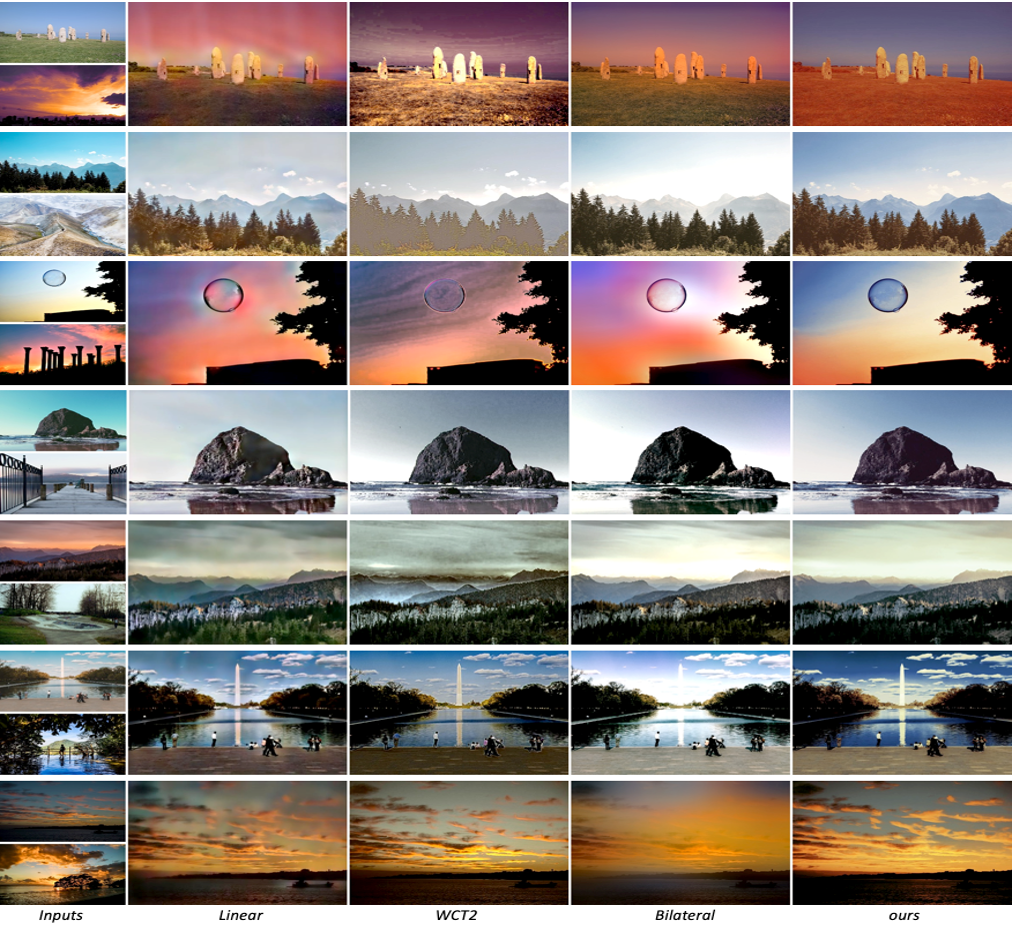}
\caption{Qualitative comparisons between our method and state-of-the-art photo-realistic stylization methods.}
\label{fig:exp_full_page}
\end{figure*}

We demonstrate a small sampling of the test set with challenging cases in Fig. \ref{fig:exp_full_page}.
Linear is mainly an artistic style transfer method with weak spatial distortions, where spatial propagation network is further applied as post-processing step to reduce remaining artifacts.
However, as shown in Fig. \ref{fig:exp_full_page}, there are still noticeable distortions, including extra blurry texture and some dirty colors.
WCT$^2$ achieves an excellent color transfer degree. However, it will generate hazy results in some cases (such as the 2-nd row). Meanwhile, when the color changes smoothly in a large region, WCT$^2$ may generate stylization results with stepped color bars (such as the 1-st, 3-rd and 7-th rows).
This phenomenon may be caused by the whiten-and-color operation, since it occurs commonly in WCT-based methods.
Bilateral performs quite well, but still has two types of artifacts. First, when the content or style image has bright light, Bilateral may generate results with  overexposure regions (such as 2-nd, 7-th, 8-th rows). Second, contrast reduction may appear in some cases, especially along strong edges. This artifact is also discussed in the original paper.
Our proposed method achieves high-quality photo-realistic stylization, where local structures of content image are well preserved without distortions. Meanwhile, the color distribution of style image is properly transferred onto content image.
In Fig. \ref{fig:exp_high}, we further demonstrate our stylization result on 4K resolution image.


\subsection{Quantitative Results}
\label{sec:exp_quantitative}

\textbf{Objective Metrics.}
As the measurement of stylization tasks are usually subjective, we adopt some objective metrics for further comparison.
Following previous works, between content images and output images, we adopt structural similarity (SSIM) index on whole images (SSIM-Whole) and their edge responses (SSIM-Edge) \cite{xie2015holistically} to measure detail preservation ability, and  use LPIPS \cite{zhang2018unreasonable} to measure perceptual similarity. 
%
%
Meanwhile, we adopt gram matrix difference (VGG style loss) between style images and output images to measure stylization effects.

All the evaluation is conducted on the dataset provided by Bilateral \cite{xia2020joint}, and the results are shown in Table. \ref{table:metric}.
%
%
Classical methods Reinhard \cite{reinhard2001color} and Piti\'{e} \cite{pitie2007linear} match the global statistics of pixels, focusing on the simple transfer like color shifts and tone curves.
So the content details of their results tend to be well preserved, namely good photorealism.
On the contrary, their gram loss is relatively high, for ignoring more complex style patterns.
Among learning-based methods, our AdaCM acheives best stylization score.
In terms of photorealism, AdaCM is slightly behind Bilateral.
We argue this is due to the training instability of the hyper-parameter network paradigm and the lack of spatial context in ColorMLP.
Besides, the photorealism scores are somewhat inconsistent with user study, because the subjective evaluation is more complex and abstract.
To sum up, these results demonstrate the effectiveness of our AdaCM.

\begin{table}[t]
\centering
\begin{center}
\begin{tabular}{cccc}
\toprule
Mean Score & Photorealism & Stylization & Overall \\ 
\midrule
Linear         & 2.79 & 2.89 & 2.69 \\
WCT$^2$     & 2.92 & 3.21 & 2.97 \\
PhotoNAS    & 2.51 & 2.59 & 2.41 \\
Bilateral   & 3.66 & 3.50 & 3.54 \\
Ours        & \textbf{3.82} & \textbf{3.51} & \textbf{3.60} \\
\bottomrule
\end{tabular}
\caption{Results of user study. Higher score is better.}
\label{table:user_study}
\end{center}
\end{table}

\noindent
\textbf{User Study.}
In this section, we conduct user study for subjective comparison. 
We recruit 30 users uncorrelated with the project.
%
Following Bilateral \cite{xia2020joint}, we randomly sample 15 content-style pairs and generate stylization results using  Linear \cite{li2018learning}, WCT$^2$ \cite{yoo2019photorealistic}, PhotoNAS \cite{an2020ultrafast}, Bilateral \cite{xia2020joint} and our AdaCM.
For each pair, we display stylized images side-by-side in  random order. And the users are asked to rate these results from 1 to 5 based on three aspects separately: photo-realistic  (fewer artifacts), stylization degree (similar to style image), and overall quality.
We totally collect 1,350 responses (15 $\times$ 30 $\times$ 3).
As shown in Table. \ref{table:user_study}, our method outperforms Linear, WCT$^2$ and PhotoNAS.
Compared to Bilateral, our method has better performance on photo-realistic, similar performance on stylization degree and slightly better performance on overall quality.

\noindent
\textbf{Efficiency Analysis.} 
We compare the efficiency of our method and other methods under multiple resolutions.
All experiments are conducted using a single Nvidia V100 GPU with 16GB RAM.
The inference time of PhotoWCT, Linear, WCT$^2$, and PhotoNAS are evaluated with official codes, and we directly refer the inference time of Bilateral \cite{xia2020joint} from the original paper since there is no open source code.
As shown in Table. \ref{table:efficiency}, our method is extremely fast and essentially invariant to practical resolutions.
Our fast inference pipeline includes three steps: parameter prediction (4.79ms), generating 3D-LUT $V_{G}$ using ColorMLP (0.26ms), and 3D-LUT based color manipulation.
Thus, similar to \cite{xia2020joint}, our method is extremely fast since the ``heavy'' encoder module of the network is fed with constant low resolution images (256 $\times$ 256).
Meanwhile, results on Table. \ref{table:efficiency} demonstrate that ColorMLP is ultrafast under low resolution, but its computation cost grows rapidly with higher resolution. 
These results also demonstrate the necessity of using 3D-LUT as an intermediary during inference.
To conclude,  our method achieves state-of-the-art photo-realistic stylization speed.

\subsection{Ablation Study}

\begin{figure}
\centering
\includegraphics[width=0.9\linewidth]{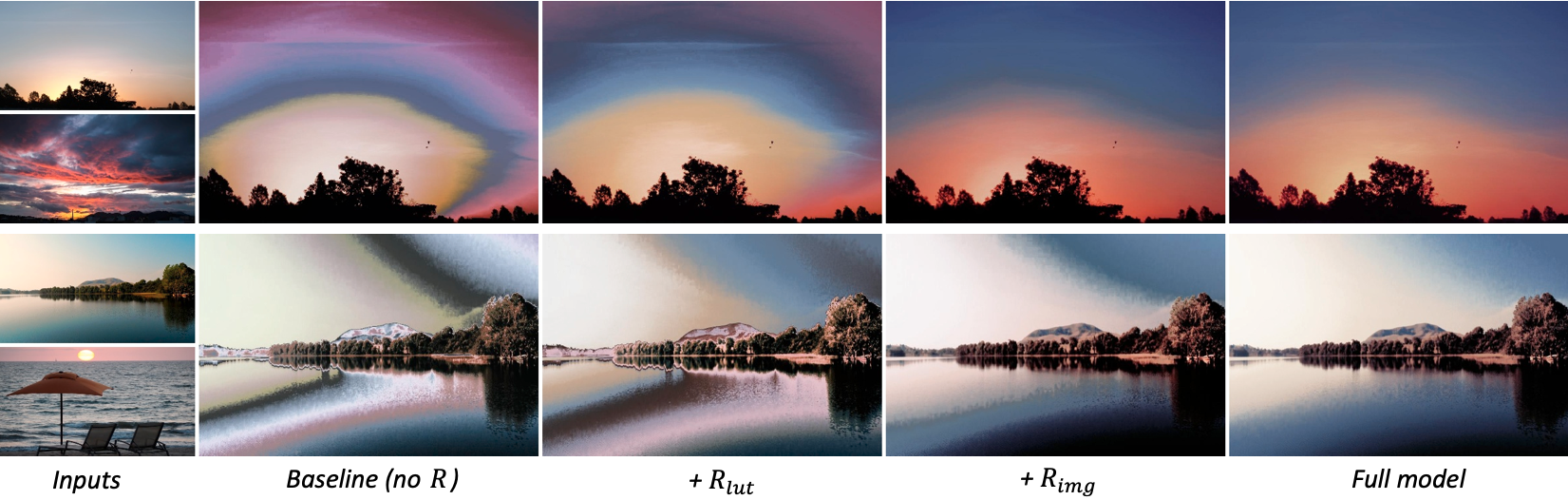}
    \caption{Ablation study of regularization term. There are no regularization terms used in baseline, and both  $\mathcal{R}_{lut}$  and $\mathcal{R}_{img}$ are used in full model. Content loss $\mathcal{L}_c$ and style loss $\mathcal{L}_c$ are adopted in all experiments.}
\label{fig:exp_ablation_loss}
\end{figure}

\textbf{Regularization.}
We conduct ablation experiments to verify the effectiveness of each regularization term used during training.
The results of two typical challenging cases are shown in Fig. \ref{fig:exp_ablation_loss}, which demonstrates that:
(1) Without using any regularization term, there are severe artifacts and distortions, especially in slow color transition region of original image such as sky. Meanwhile, the relative brightness relation in original image is broken, resulting in unnatural color distribution;
(2) Adding regularization $\mathcal{R}_{lut}$ on 3D-LUT, artifacts are partially removed but still obvious. This result indicates that adding smoothness regularization uniformly on full color space is insufficient;
(3) Adding regularization $\mathcal{R}_{img}$ only on stylization image, artifacts are largely removed and the relative brightness relation is well preserved. However, there are still artifacts in sky region;
(4) Adding $\mathcal{R}_{lut}$ and $\mathcal{R}_{img}$ simultaneously, our full model can remove nearly all artifacts and generate photo-realistic stylization results. The final results indicate that $\mathcal{R}_{lut}$ and $\mathcal{R}_{img}$ are complementary during training AdaCM.

\begin{figure}[t]
\centering
\includegraphics[width=0.85\linewidth]{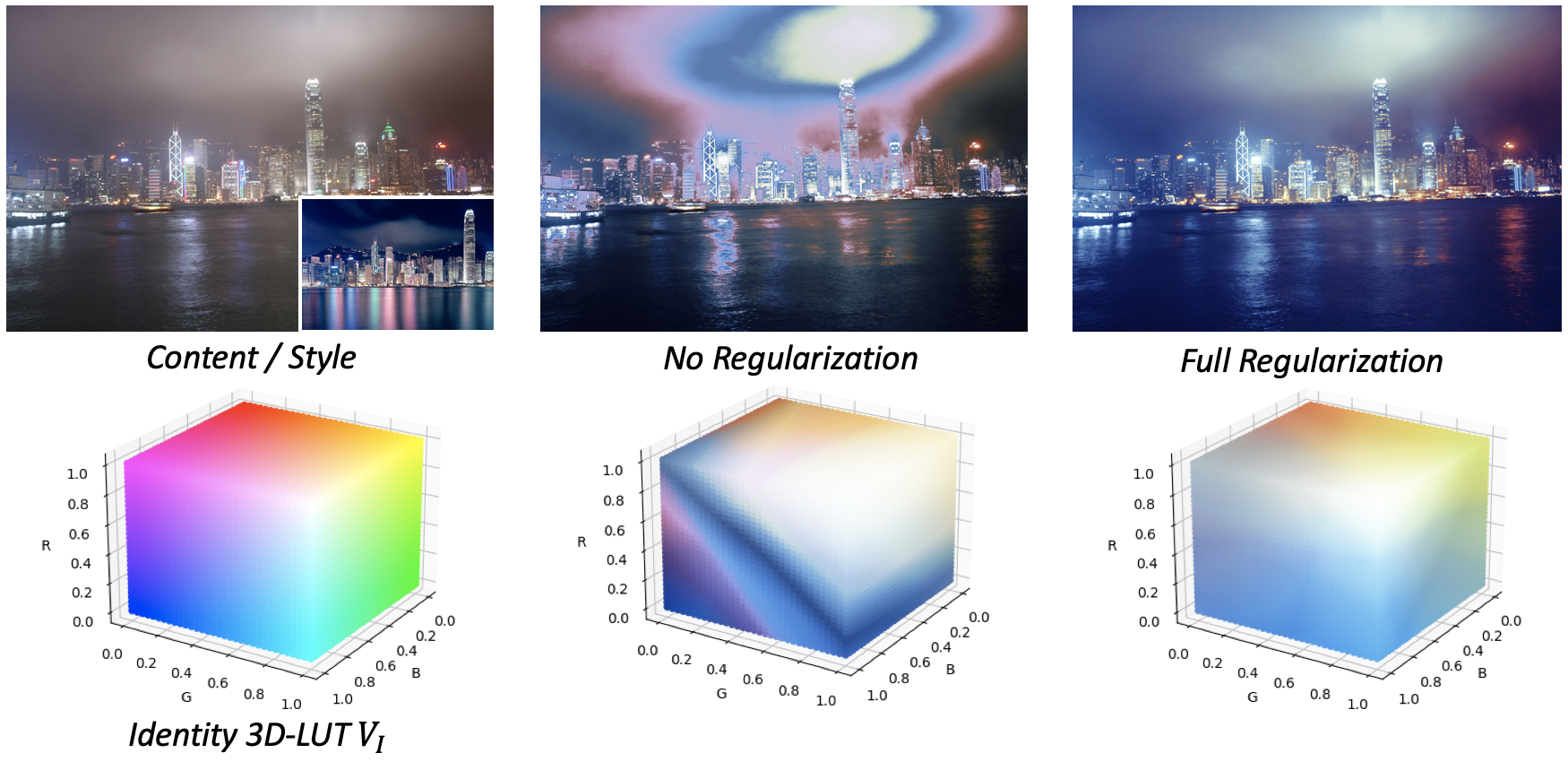}
   \caption{Visualization of generated ColorMLP with the help of 3D-LUT. Results of models trained with and without regularization term are both illustrated. }
\label{fig:exp_ablation_lut}
\end{figure}

To further evaluate the effect of regularization, we visualize the color mapping relationship in ColorMLP with the help of 3D-LUT.
As shown in Fig. \ref{fig:exp_ablation_lut}, if no regularization used, there are severe artifacts. Meanwhile, via generated 3D-LUT, we can find there exist drastic changes in color space, along with the broken of brightness monotonicity. With regularization, these artifacts are removed and we can obtain a smooth color mapping space.

\begin{table}
\centering
    \begin{tabular}{lm{1.0cm}<{\centering}m{1.0cm}<{\centering}m{1.0cm}<{\centering}m{1.0cm}<{\centering}}
        \toprule
        \multirow{2}{1.6cm}{Method} & \multicolumn{4}{c}{Inference Time Per Image} \\
        \cline{2-5}
        & $512^2$ & $1024^2$ & $2000^2$ & $4000^2$\\
        \midrule
        PhotoWCT  & 3.13s & 3.32s & OOM & OOM\\
        LST  & 1.32s & 2.16s & OOM & OOM\\
        WCT$^2$ & 5.64s & 9.41s & OOM & OOM \\
        PhotoNAS & 0.39s & 0.60s & OOM & OOM \\
        Bilateral* & $<$5ms & $<$5ms & $<$5ms & $<$5ms\\
        \hline
        ColorMLP  & 0.42ms & 1.64ms & 5.93ms & 30.9ms \\
        3D-LUT & 0.06ms & 0.09ms & 0.32ms & 1.09ms \\
        \hline
        Ours & 5.11ms & 5.14ms & 5.37ms & 6.14ms \\
        \bottomrule
    \end{tabular}
    \caption{Efficiency comparison under multiple resolutions on an Nvidia V100 GPU (16GB RAM). OOM indicates out-of-memory. * indicates the speed  refers to its original paper. We evaluate the speed of ColorMLP with $N=20$ and 3D-LUT with $M=33$. For AdaCM, the fast color manipulation strategy is used. }
    \label{table:efficiency}
\end{table}

\begin{table}
\centering
\begin{center}
\scalebox{0.85}{\begin{tabular}{cccccc}
\toprule
$M$ of 3D-LUT & 8 & 16 & 33 & 64 & 128 \\ 
\midrule
PSNR & 36.36 & 41.64 & 45.82 & 49.25 & 53.30 \\
\bottomrule
\end{tabular}}
\caption{Ablation study on the approximate ability of 3D-LUT with different $M$.}
\label{table:lut_size}
\end{center}
\end{table}

\begin{figure}[t]
\centering
\includegraphics[width=0.90\linewidth]{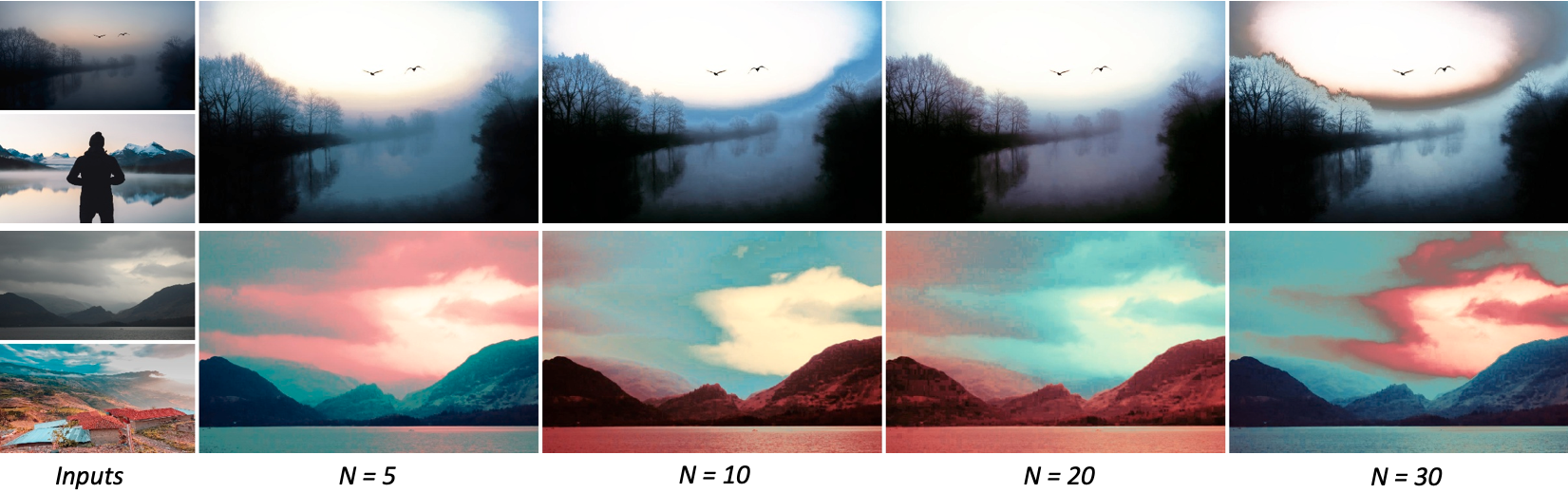}
  \caption{Ablation study of effects of number of hidden units used in ColorMLP.}
\label{fig:exp_ablation_units}
\end{figure}

\noindent
\textbf{Number of Hidden Units.}
In Sec. 3D-LUT, we have discussed how many hidden units are enough for fitting a general 3D-LUT.
In this section, we further conduct ablation experiments to verify how many hidden units are proper for photo-realistic stylization task.
We train four AdaCM models with $N=5, 10, 20, 30$ separately, and show two challenging cases in Fig. \ref{fig:exp_ablation_units}, which demonstrate that:
(1) with small number of hidden units (i.e. $N=5$), AdaCM can only model relatively simple color mapping relationship due to its limited capacity;
(2) with large number of hidden units (i.e. $N=30$), there appears some severe artifacts in stylization results. We believe the main reason is that directly predicting large number of ColorMLP parameters is too difficult for the encoder.   
Based on these results, we find $N=20$ is a proper number for hidden units, which can generate robust stylization results. 

\noindent
\textbf{Number of Bins.}
To improve inference efficiency in AdaCM, we design a fast pipeline with the help of 3D-LUT.
The precision of transformation is determined by the number of bins $M$. 3D-LUT with larger $M$ can achieve better approximation of ColorMLP.
We conduct further ablation experiments to choose the proper value of $M$.
We compute the average peak signal to noise ratio (PSNR) between stylized images directly generated by ColorMLP and generated by 3D-LUTs with different $M$ on the evaluation dataset.
As shown in Table. \ref{table:lut_size}, PSNR becomes higher with the increasing of $M$.
When $M=33$, PSNR reaches 45.82, which is a high score and is hard for human to distinguish the differences. 
Meanwhile, $M=33$ is also a common choice empirically.
Thus, 33 bins is a proper value in the fast inference pipeline.

\section{Conclusions}

In conclusion, we propose a novel photo-realistic style transfer method AdaCM, which can synthesize high-resolution stylized image efficiently and effectively.
In this work, we reveal the powerful ability of a lightweight MLP model to approximate complex non-linear color mapping relation. 
Then we utilize this remarkable property to formulate the photo-realistic stylization task as a simple MLP parameters prediction task.
Experiments demonstrate that our method can generate high-quality and stable stylization results. Meanwhile, our methods can achieve real-time inference in 4K-resolution images.
We believe the fast inference speed and the simple implementation of our method can enable more style transfer applications, especially in mobile devices.
As for future work, we hope to further simplify the network structure to achieve faster speed, and explore better training constraint to achieve better stylization  quality.

\bibliography{aaai23}

\clearpage

\begin{appendices}

\noindent\textbf{\huge Appendix}

\section{Details of Parameter Prediction Module}
As stated in the main paper, the Parameter Prediction Module (PPM) is designed to model the color mapping relationship between content and style images via predicting the parameters of ColorMLP. 

Following \cite{xia2020joint}, we first adopt a pretrain VGG-19 \cite{simonyan2015very} network to extract feature maps of content and style images at four scales (conv1\_1, conv2\_1, conv3\_1 and conv4\_1). Then we use three \emph{Style-based Splatting} \cite{xia2020joint} blocks to fuse AdaIN-aligned \cite{huang2017arbitrary} feature maps in multiple granularity. Each \emph{Style-based Splatting} block contains a stride-2 weight-sharing convolutional layer to learn the joint distribution between content and style features, as well as a stride-1 convolutional layer to integrate learned and pretrain AdaIN-aligned features. At last, we combine local context and global scene summary to predict parameters $\Theta$ of ColorMLP via multiple convolution and fully-connected layers. Table. \ref{tab:arch} provides the details of our Parameter Prediction Module.

\section{Additional Results}

\subsection{Qualitative Results}
\begin{figure*}[t]
\centering
\includegraphics[width=1.0\textwidth]{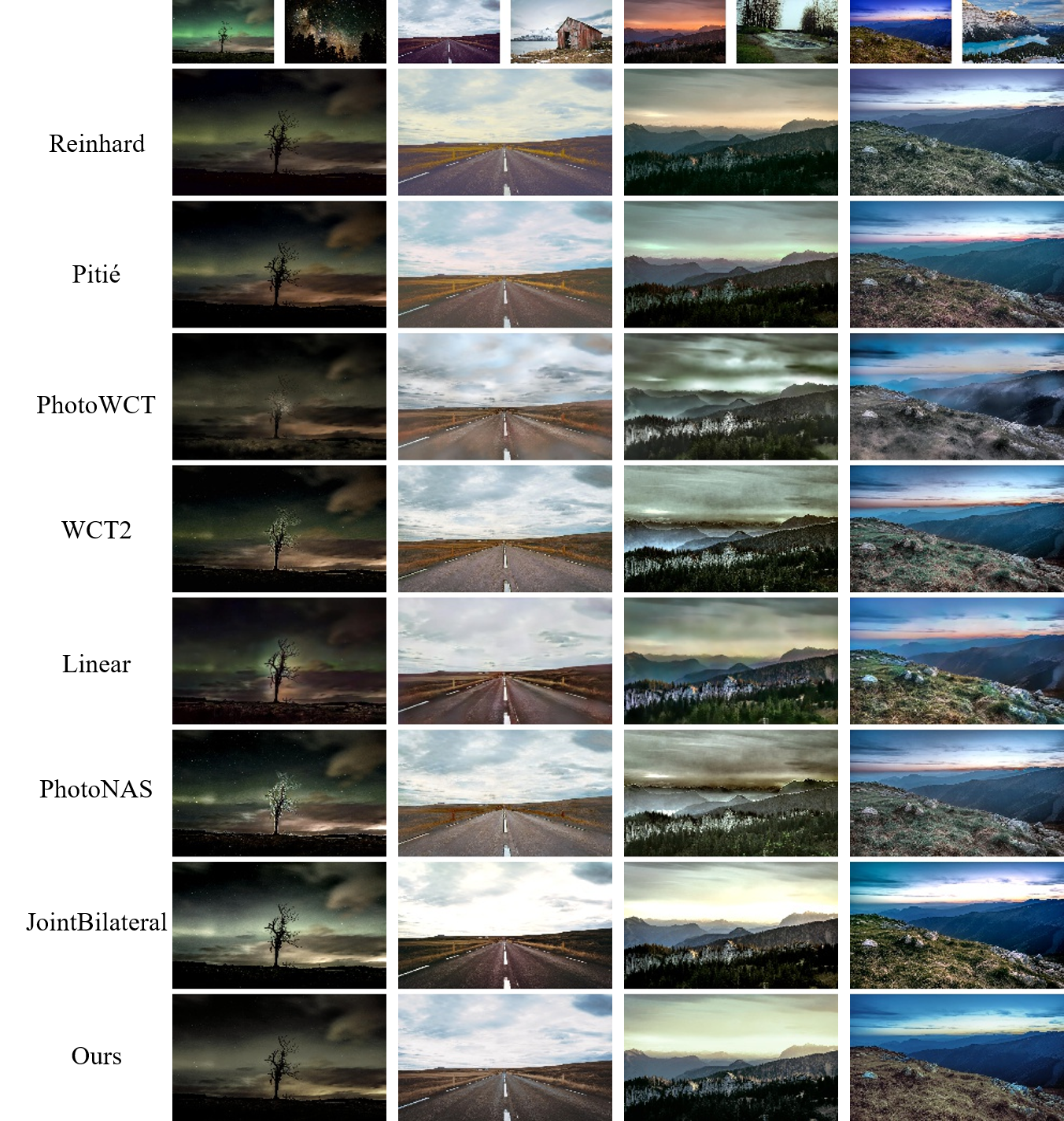}
\caption{Additional qualitative results against classical methods and other state-of-the-art photo-realistic stylization methods.}
\label{fig:qua1}
\end{figure*}

\begin{figure*}[t]
\centering
\includegraphics[width=1.0\textwidth]{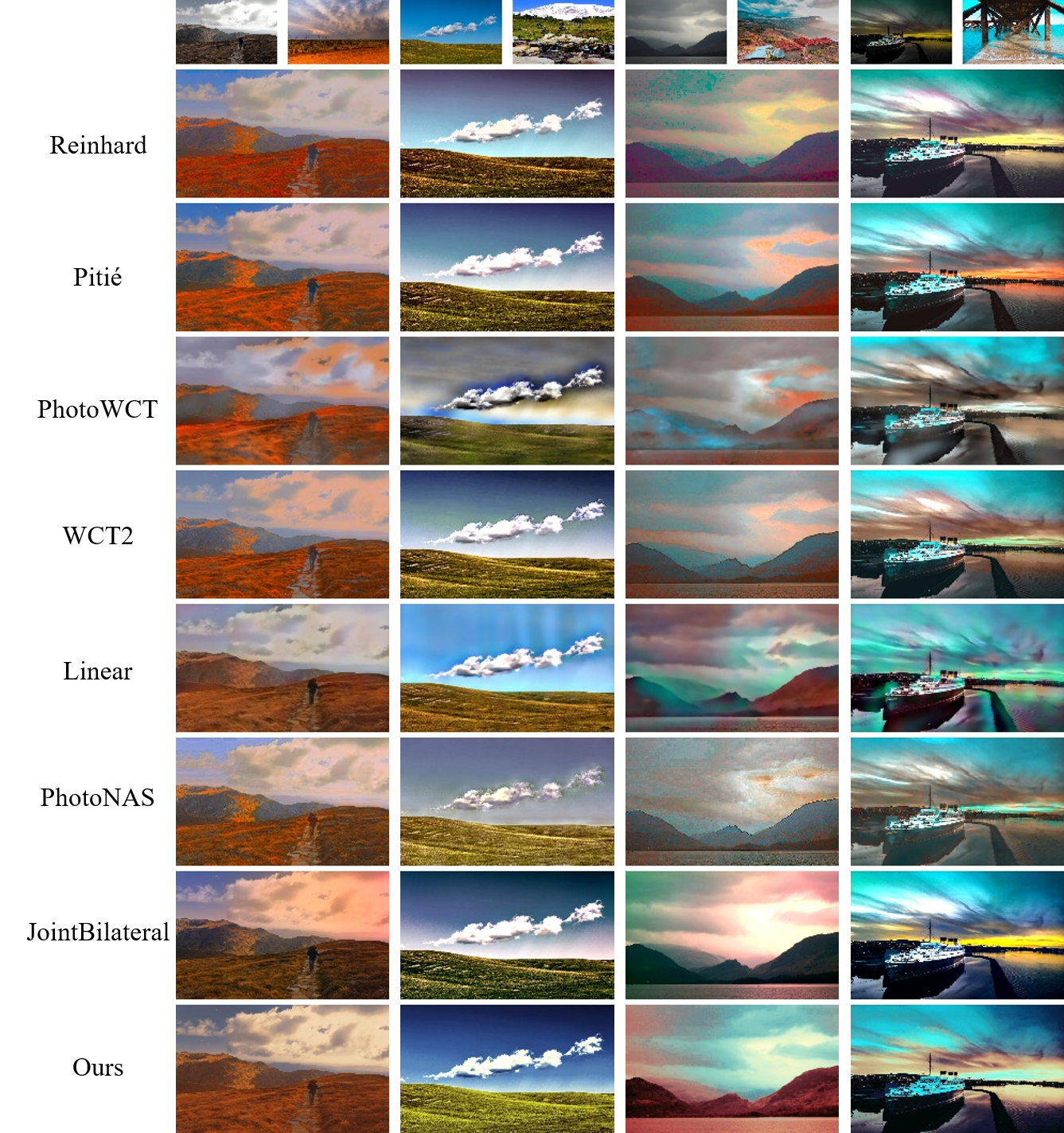}
\caption{Additional qualitative results against classical methods and other state-of-the-art photo-realistic stylization methods.}
\label{fig:qua2}
\end{figure*}

We provide more qualitative results against classical non-neural methods: Reinhard \cite{reinhard2001color}, Piti\'{e} \cite{pitie2007linear} and other learning-based photo-realistic style transfer methods: Linear \cite{li2018learning}, WCT$^2$ \cite{yoo2019photorealistic}, Bilateral \cite{xia2020joint}, PhotoWCT \cite{li2018closed} and PhotoNAS \cite{an2020ultrafast} using default settings.
%
%
As shown in Fig. \ref{fig:qua1}, and Fig. \ref{fig:qua2}, when input style patterns are rich, classical methods can not transfer them reasonably, resulting in inconsistent color tone between output images and style images (especially Reinhard \cite{reinhard2001color}).
Our AdaCM achieves high-quality photo-realistic stylization results, while other learning-based methods suffer from severe distortions, blurry texture, stepped color bars and overexposure in some cases. 
It's worth noting that our method produces reasonable results without any post-processing step or work with the assist of hard-to-obtain semantic masks.

\begin{table}[t]
    \centering
    \scalebox{0.95}{
    \begin{tabular}{c|c c c c}
    \hline
        layer & type & stride & spatial size & channels\\
        \hline
        \emph{Splatting}$_1^1$  & \emph{conv}3x3 & 2 & 128 & 32 \\
        \emph{Splatting}$_1^2$  & \emph{conv}3x3 & 1 & 128 & 32 \\
        \emph{Splatting}$_2^1$  & \emph{conv}3x3 & 2 & 64 & 64 \\
        \emph{Splatting}$_2^2$  & \emph{conv}3x3 & 1 & 64 & 64 \\
        \emph{Splatting}$_3^1$  & \emph{conv}3x3 & 2 & 32 & 128 \\
        \emph{Splatting}$_3^2$  & \emph{conv}3x3 & 1 & 32 & 128 \\
        \hline
        \emph{Fusion}$^1$  & \emph{conv}3x3 & 2 & 16 & 128 \\
        \emph{Fusion}$^2$  & \emph{conv}3x3 & 1 & 16 & 128 \\
        \hline
        \emph{Local}$^1$  & \emph{conv}3x3 & 1 & 16 & 256 \\
        \emph{Local}$^2$  & \emph{conv}3x3 & 1 & 16 & 256 \\
        \hline
        \emph{Global}$^1$  & \emph{conv}3x3 & 2 & 8 & 128 \\
        \emph{Global}$^2$  & \emph{conv}3x3 & 2 & 4 & 128 \\
        \emph{Global}$^3$  & \emph{conv}3x3 & 2 & 2 & 128 \\
        \emph{Global}$^4$  & \emph{fc} & - & - & 256 \\
        \emph{Global}$^5$  & \emph{fc} & - & - & 256 \\
        \hline
        \emph{Fusion}$^3$  & \emph{conv}1x1 & 1 & 16 & 256 \\
        \emph{Fusion}$^4$  & \emph{conv}3x1 & 1 & 16 & 256 \\
        \hline
        \emph{GAP} & \emph{pooling} & - & 1 & 256 \\
        \emph{P}  & \emph{fc} & - & - & 563 \\
        \hline 
    \end{tabular}}
    \caption{\small Details of our PPM. The pipeline mainly follows \cite{xia2020joint}. \emph{Splatting}$_j^i$ denotes the \emph{i-th} layer in the \emph{j-th} \emph{Splatting} block. We apply AdaIN after each \emph{Splatting}$_j^1$. \emph{Local}, \emph{Global} and \emph{Fusion} refer to local branch, global branch and fusion layer respectively. We acquire feature summary via \emph{Global Average Pooling} after the last fusion layer, and use it to predict parameter of ColorMLP.}
    \label{tab:arch}
\end{table}

\subsection{Content-dependent Mapping}
\begin{figure*}[t]
\centering
\includegraphics[width=0.9\textwidth]{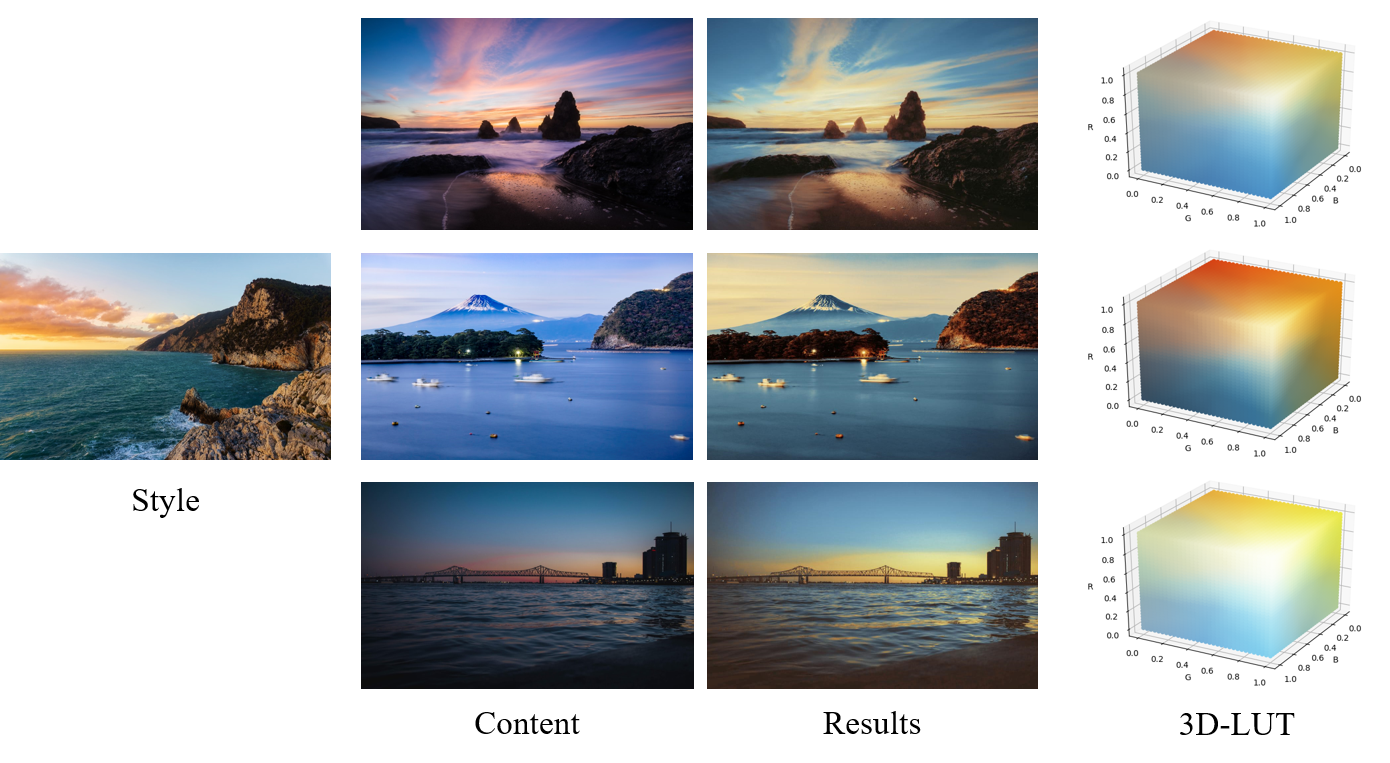}
\caption{Qualitative results of same style image and different content images.}
\label{fig:trans}
\end{figure*}

Although the input of ColorMLP is the RGB color, AdaCM has implicitly encoded local and global context into the parameters of ColorMLP by PPM.
To verify the color mapping learned by AdaCM is content-dependent, we use the same style image and different content images to get stylized results and visualize color mappings via 3D-LUT.
As is shown in Fig. \ref{fig:trans}, the learned color mapping changes with the content image, demonstrating that our method does not produce results through context-free statistics matching like classical methods.

\subsection{Perceptual Loss}
\begin{figure*}[t]
\centering
\includegraphics[width=0.85\textwidth]{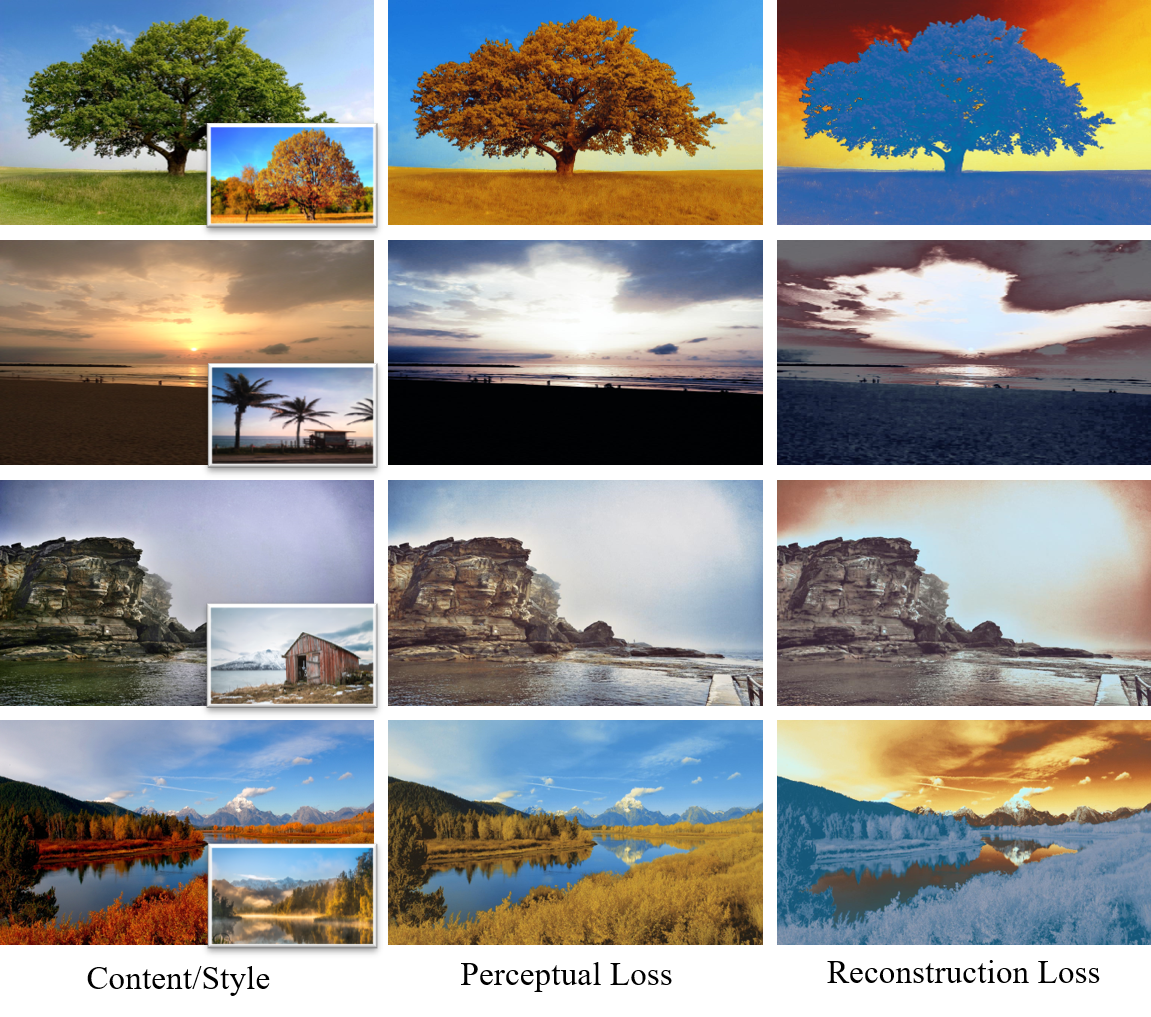}
\caption{Qualitative comparisons between perceptual loss and reconstruction loss.}
\label{fig:percep}
\end{figure*}

Perceptual loss is the most basic loss function in style transfer and other generative tasks.
It helps maintain the semantic content of the image during stylization.
We conduct another experiments to verify its effectiveness by replacing it with pixel-level mean square loss (reconstruction loss).
As shown in Fig. \ref{fig:percep}, AdaCM would learn incorrect transform without the constraint of perceptual loss, resulting in distorted and unreasonable images.

\subsection{Fitting experiment of ColorMLP}
\begin{figure*}[t]
\centering
\includegraphics[width=0.90\textwidth]{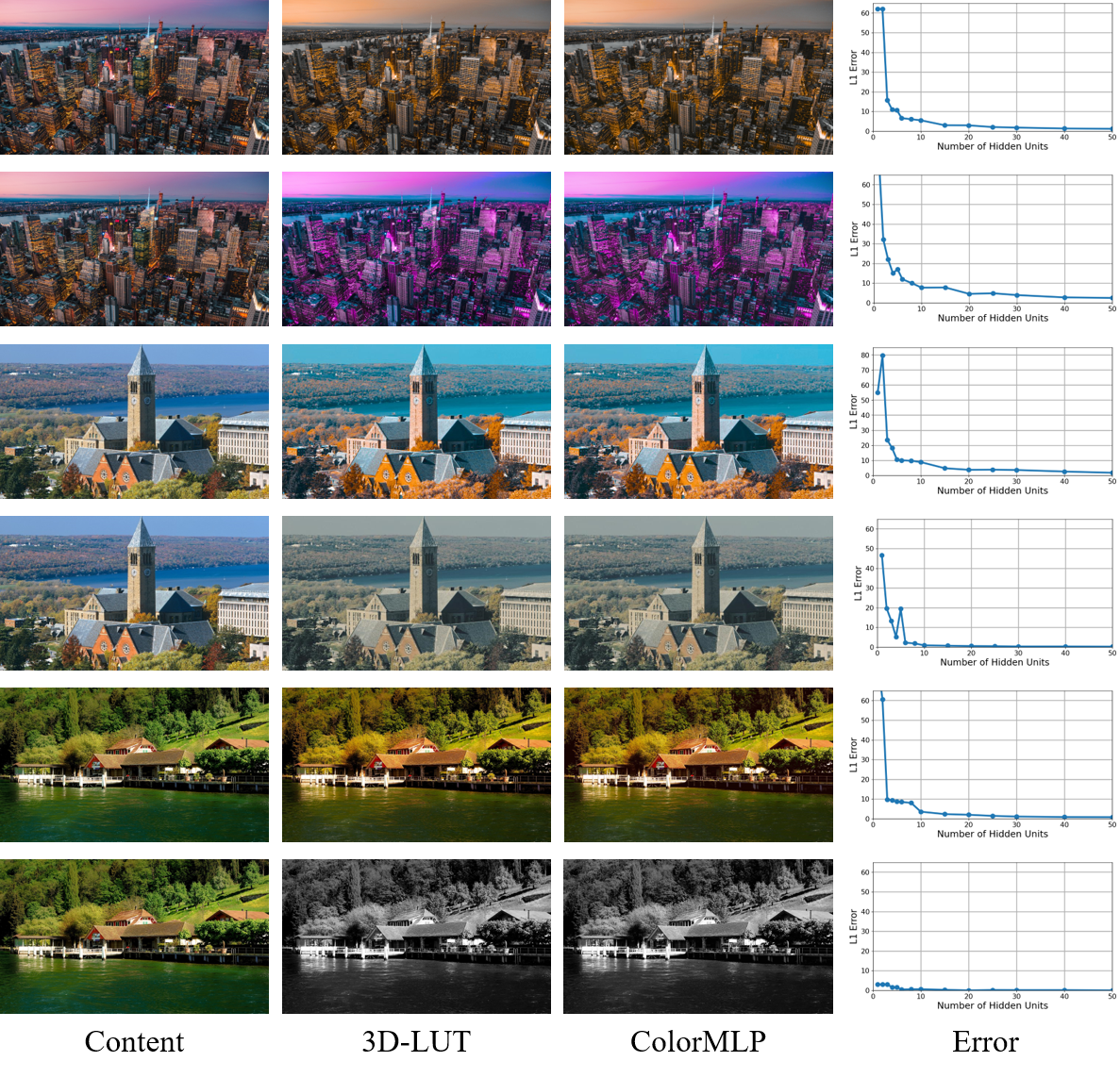}
\caption{Visualization of the ColorMLP fitting experiment. (1) The first column illustrates the input images. (2) The second column illustrates the images processed by different 3D-LUTs. (3) We train ColorMLP (N=20) to fit 3D-LUT in each row, and process input images with corresponding trained ColorMLP. The resutls are shown in third column. (4) In the last column, for each 3D-LUT and ColorMLP, we demonstrate the relationship between L1 error (after training) and number of hidden units.}
\label{fig:lut}
\end{figure*}

We provide more detailed results of our ColorMLP fitting experiment (Sec. 3.1 of main paper). 
We totally collect 10 3D-LUTs from the internet and train ColorMLPs to fit each 3D-LUT. 
In this section, we choose six typical 3D-LUTs (\emph{Blackgold}, \emph{Cyberpunk}, \emph{GreenOrange}, \emph{Muted}, \emph{Aragonite} and \emph{Bloodstone}) and visualize experiment results.
In Fig. \ref{fig:lut}, we demonstrate images processed by 3D-LUTs and trained ColorMLPs (N=20) in second and third columns respectively. These results demonstrate that trained ColorMLP can achieve indistinguishable results compared to 3D-LUT.
Meanwhile, for each 3D-LUT, we also demonstrate the fitting results of \emph{average L1 error} versus \emph{hidden units number} in the last column.

\subsection{Video Stylization}
\begin{figure*}[t]
\centering
\includegraphics[width=0.95\textwidth]{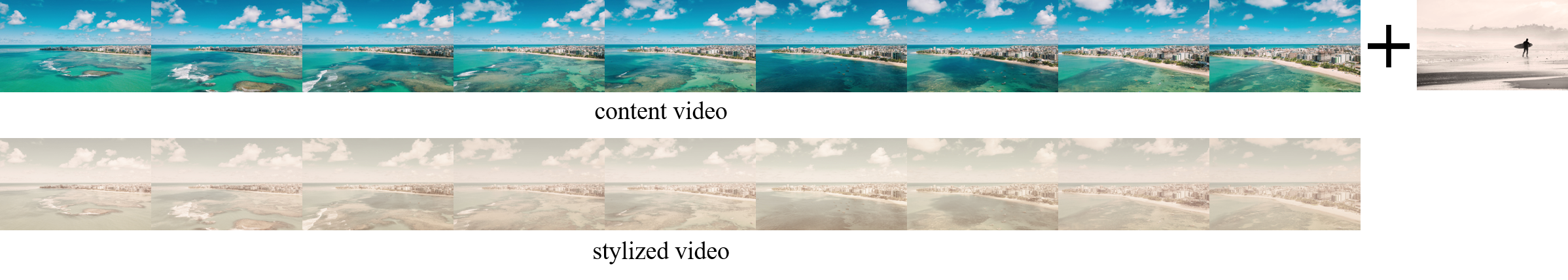}
\caption{Photo-realistic video stylization results. Our method can produce temporally coherent frames.}
\label{fig:video}
\end{figure*}

Since our proposed AdaCM has a strong ability to maintain the smoothness of semantically consistent regions, it can generate temporally coherent video results without additional constraint likes optical flow. In Fig. \ref{fig:video} , we show a sequence of stylized frames. The style is consistent, and the transferred video is quite stable.

\end{appendices}

\end{document}